%% file: Li.CVPR.2025.tex
\documentclass[10pt,twocolumn,letterpaper]{article}

\usepackage[pagenumbers]{cvpr} 

\input{section/preamble}

\definecolor{cvprblue}{rgb}{0.21,0.49,0.74}
\usepackage[pagebackref,breaklinks,colorlinks,allcolors=cvprblue]{hyperref}
\usepackage{url}
\usepackage{bbm}
\usepackage{amsmath}
\usepackage{amssymb}
\usepackage{multirow}

\def\paperID{7580} 
\def\confName{CVPR}
\def\confYear{2025}

\title{Any2Any: Incomplete Multimodal Retrieval with Conformal Prediction}

\author{Po-han Li\textsuperscript{1} 
\qquad
Yunhao Yang\textsuperscript{1}
\qquad
Mohammad Omama\textsuperscript{1}
\qquad
Sandeep Chinchali\textsuperscript{1} 
\qquad
Ufuk Topcu\textsuperscript{1}
\\
$^1$UT Austin
}

\begin{document}
\maketitle

\begin{abstract}
\input{section/abstract}
\end{abstract}

\input{section/intro}
\input{section/related_works}
\input{section/problem_formulation}
\input{section/preliminary}
\input{section/method}
\input{section/experiments}
\input{section/conclusion}

\clearpage
{
    \small
    \bibliographystyle{ieeenat_fullname}
    \bibliography{bib/external, bib/swarm}
}

\clearpage
\appendix
\input{section/appendix}


\end{document}

%% file: section/preamble.tex
\newcommand{\Figref}[1]{Fig.~\ref{#1}}
\newcommand{\Tabref}[1]{Table~\ref{#1}}
\newcommand{\Eqref}[1]{Eq.~\ref{#1}}
\newcommand{\Secref}[1]{Sec.~\ref{#1}}
\newcommand{\Appref}[1]{Appx.~\ref{#1}}
\newcommand{\argmin}{\arg\min}
\newcommand{\argmax}{\arg\max}
\newcommand{\ceil}[1]{\left\lceil #1 \right\rceil}

\newcommand{\data}[2]{x_{#1}^{#2}}
\newcommand{\qdata}[2]{\chi_{#1}^{#2}}
\newcommand{\dataset}[1]{X^{#1}}
\newcommand{\obsdataset}[1]{\hat{X}^{#1}}
\newcommand{\obsdata}[1]{\hat{x}_{#1}}
\newcommand{\obsqdata}[1]{\hat{\chi}_{#1}}

\newcommand{\dpnum}{N^{\mathrm{r}}}
\newcommand{\qnum}{N^{\mathrm{q}}}
\newcommand{\modnum}{n}
\newcommand{\qmodnum}{\eta}
\newcommand{\calinum}{\nu}
\newcommand{\mattonum}{F}

\newcommand{\newsim}{\mathcal{S}}
\newcommand{\cossim}{\mathcal{S}_{\mathrm{ori}}}
\newcommand{\bisim}{\mathcal{S}_{\mathrm{cross}}}
\newcommand{\simmat}[2]{\mathbb{S}_{#1 #2}}
\newcommand{\simscore}{\theta}

\newcommand{\caliset}{\Phi}

\newcommand{\caliscore}{\gamma}
\newcommand{\predband}{\hat{C}}
\newcommand{\conmat}[2]{\mathbb{C}_{#1 #2}}

\newcommand{\zero}{\mathbf{0}}

\newcommand{\modal}[1]{m^{#1}}
\newcommand{\qmodal}[1]{\mu^{#1}}


%% file: section/abstract.tex
Autonomous agents perceive and interpret their surroundings by integrating multimodal inputs, such as vision, audio, and LiDAR.
These perceptual modalities support retrieval tasks, such as place recognition in robotics.
However, current multimodal retrieval systems encounter difficulties when parts of the data are missing due to sensor failures or inaccessibility, such as silent videos or LiDAR scans lacking RGB information.
We propose \textbf{Any2Any}–a novel retrieval framework that addresses scenarios where both query and reference instances have incomplete modalities.
Unlike previous methods limited to the imputation of two modalities, Any2Any handles any number of modalities without training generative models.
It calculates pairwise similarities with cross-modal encoders and employs a two-stage calibration process with conformal prediction to align the similarities. 
Any2Any enables effective retrieval across multimodal datasets, \textit{e.g.}, text-LiDAR and text-time series.
It achieves a Recall@$5$ of $35\%$ on the KITTI dataset,  which is on par with baseline models with complete modalities.

%% file: section/intro.tex
\section{Introduction}
Autonomous agents perceive the world through multiple senses, integrating various modalities to understand the surroundings.
Recent advances in multimodal machine learning have created new opportunities for retrieval \cite{wang2016comprehensivesurveycrossmodalretrieval} but face limitations due to incomplete data from sensor failures or data inaccessibility \cite{shi2024incomplete, jing2020incomplete}.
For example, autonomous vehicles use LiDAR and RGB cameras for place recognition, a retrieval task that identifies the agents' locations.
However, it is challenging in areas with limited sensor coverage due to private property restrictions or varying light conditions.
Also, due to legal or technical issues, existing text-to-video retrieval datasets lack audio or visual information in some instances \cite{xu2016msr}, and the same applies to other domains like time series \cite{bitcoin_dataset}.
In this paper, we focus on multimodal retrieval scenarios where both query and reference instances have incomplete modalities, such as silent videos or LiDAR scans missing RGB information.

Previous works also examined scenarios of incomplete multimodal instances in retrieval systems \cite{shi2024incomplete, jing2020incomplete, li2022general}.  
However, they are limited to bimodal retrieval and necessitate the training of generative imputation models.
Moreover, these methods face scalability challenges as the number of modalities grows and struggle with new modality pairs, such as text-LiDAR, due to limited training data for imputation models.

\input{figure_latex/any2any_system}
We thus propose \textit{Any2Any}, a novel retrieval framework that enables retrieval from \textit{Any} query modality \textit{to} \textit{Any} reference modality.
This framework facilitates multimodal data retrieval in scenarios where instances contain incomplete modalities without training any models.
It employs cross-modal encoders to process existing modalities of instances and calculate their pair-wise cross-modal similarities, as shown in the left of \Figref{fig:any2any_system}. 
However, as we later examine, the pair-wise cross-modal similarities are not directly comparable because the encoder outputs have different ranges.
It thus employs a two-stage calibration process. The first stage utilizes conformal prediction to standardize and align similarity scores to probabilities of correct retrieval.
The second stage converts multiple probabilities into a scalar that represents the overall probabilities of correct retrieval across all modality pairs.
Any2Any enables direct comparisons between query and reference instances, even when they have different and incomplete modalities.
Using conformal prediction, it is compatible with any number and combination of modalities, enabling it to adapt to any multimodal datasets, such as KITTI \cite{KITTI} (vision, LiDAR, text), MSR-VTT \cite{xu2016msr} (text, audio, vision), and Monash Bitcoin \cite{bitcoin_dataset} (time series, text).

\textbf{Contributions.}
Our contributions are threefold:
(1) We introduce Any2Any, a multimodal retrieval framework capable of retrieving instances with incomplete modalities, which requires no training of models.
(2) We extended the definition of conformal prediction to vector-based retrieval for calibration. We framed the retrieval task as a binary classification problem. It predicts whether the retrieval is correct from the similarity scores (\Secref{sec:prelim}).
(3) We verified the capability of the Any2Any framework on $3$ real-world datasets, KITTI \cite{KITTI}, MSR-VTT \cite{xu2016msr}, and Monash Bitcoin \cite{bitcoin_dataset}.
The datasets include challenging yet unexplored modality pairs such as text-LiDAR and text-time series (\Secref{sec:exps}), paving the way for novel algorithms and insights in multimodal retrieval.
Even with incomplete information, the experimental results show that Any2Any's performance is on par with baselines with complete modalities.
We also evaluate the generalization ability of the Any2Any framework across various cross-modal encoders and analyze the effectiveness of these encoders on the overall retrieval performance.

%% file: figure_latex/any2any_system.tex
\begin{figure*}[t!]
  \centering
  \includegraphics[width=0.95\textwidth]{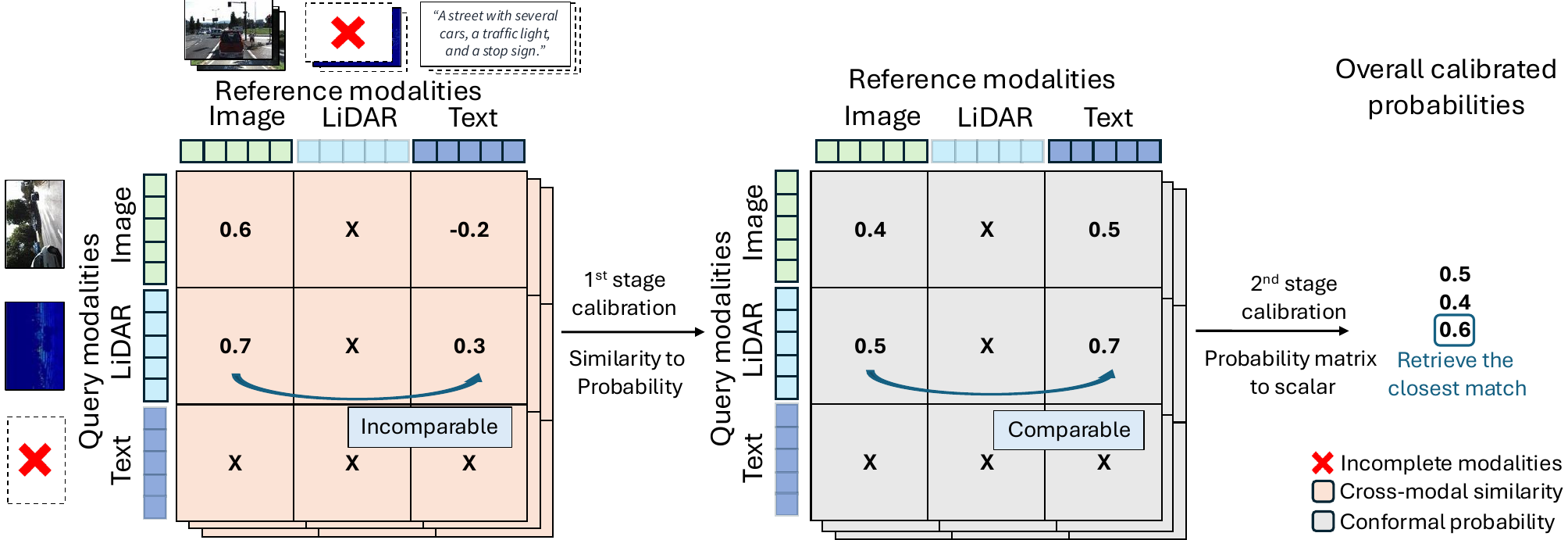}
  \caption{\small{\textbf{The Any2Any retrieval framework} retrieves multimodal data with varying incomplete modalities across instances.
    We employ a two-stage calibration process using conformal prediction to facilitate comparisons between query and reference instances, each has different incomplete modalities.
    Any2Any supports any number and combination of modalities, enabling it to adapt to any multimodal retrieval dataset.
    This illustrative figure uses data from the KITTI dataset \cite{KITTI} and captions generated by LLaVA \cite{liu2023improvedllava}.
    }}
  \label{fig:any2any_system}
\end{figure*}

%% file: section/related_works.tex
\section{Related Works}

\subsection{Cross-modal Similarity Scores}
Cross-modal similarity scores are crucial in evaluating the similarity between multimodal data, such as image-text, audio-text, and lidar-image.
CLIP (Contrastive Language-Image Pretraining) encoders \cite{clip, OpenCLIP} have been instrumental in this domain. These encoders encode images and text into a shared embedding space, enabling the computation of similarity scores using cosine similarity.
Inspired by CLIP, other works also proposed encoders for other modalities. For instance, CLAP \cite{laionclap2023} and AudioCLIP \cite{Audioclip} encode audio and text, and Liploc \cite{shubodh2024liploc} encodes lidar and RGB images.
ImageBind \cite{girdhar2023imagebind} encodes $6$ modalities–images, text, audio, inertial measurement units, heat maps, and depth–into a unified embedding space.
However, not all cross-modal similarity scores have corresponding encoders. For example, there are no lidar-text encoders for retrieval, only lidar language models \cite{yang2023lidar}.
CSA \cite{li2024csa} offers a data-efficient method to encode any bimodal data into a shared space using two unimodal encoders. It thus enables the calculation of similarity scores between any two modalities with the support of unimodal encoders, which enables our Any2Any retrieval.

\subsection{Cross-modal Retrieval}
Cross-modal retrieval is a technique enabling the search and retrieval of information across different modalities, such as text, images, and audio \cite{wang2016comprehensivesurveycrossmodalretrieval}. 
It has attracted significant attention in fields such as computer vision, natural language processing, and speech processing.
Most existing works are limited to bimodal retrievals, such as image-to-text \cite{chen2024internvl}, text-to-audio \cite{AudioRetrieval}, and image-to-lidar retrieval \cite{shubodh2024liploc}.
Some extend to three modalities, such as text-to-video (audio and image) retrieval \cite{fang2021clip2videomasteringvideotextretrieval}. 
In this paper, we address retrieval across multiple modalities and expand existing bimodal retrieval datasets to include additional modalities, showcasing the capabilities of the proposed Any2Any framework.

\subsection{Calibration in Machine Learning}
Calibration in machine learning \cite{pmlr-v70-guo17a} involves adjusting models' output probabilities to reflect the actual realizations.
Well-calibrated models provide reliable confidence levels, crucial for estimating uncertainties in fields like medical diagnoses \cite{kononenko2001machine} and autonomous driving \cite{yang2024uncertainty}.
Typical calibration techniques include Platt scaling \cite{platt1999probabilistic}, temperature scaling \cite{pmlr-v70-guo17a}, Bayesian binning \cite{naeini2015obtaining}, and conformal prediction \citep{conformal-prediction,shafer2008tutorial}.
All calibration techniques are compatible with the proposed framework. However, we focus on \textit{conformal prediction} due to its theoretical guarantees.
It generates statistically valid prediction sets of outcomes, including sets of potential outcomes for classification tasks.
Previous work \cite{xu2024conformalrankedretrieval} has extended conformal prediction to ranked retrieval systems, while we extend the definition of conformal prediction to vector-based retrieval. 

%% file: section/problem_formulation.tex
\section{Problem Formulation–Incomplete Multimodal Retrieval}
We now formally define the incomplete multimodal retrieval problem.
We first define a multimodal reference set with $\dpnum$ instances as $\dataset{\mathrm{r}} = \{ \data{1}{}, \data{2}{},..., \data{\dpnum}{} \}$, where each instance $\data{i}{} = (\data{i}{\modal{1}}, \data{i}{\modal{2}},..., \data{i}{\modal{\modnum}})$ consists of $\modnum$ modalities.
We define the $\modal{j}$ modal component of $\data{i}{}$ as $\data{i}{\modal{j}}$, with each component represented as a vector in a distinct dimension.
Similarly, we have a query set with $\qnum$ instances as $\dataset{\mathrm{q}} = \{ \qdata{1}{}, \qdata{2}{},..., \qdata{\qnum}{} \}$, where each instance $\qdata{i}{} = (\qdata{i}{\qmodal{1}}, \qdata{i}{\qmodal{2}},..., \qdata{i}{\qmodal{\qmodnum}})$ consists of $\qmodnum$ modalities. Note that $\qmodnum$ need not to be equal to $\modnum$. The query set can have different modalities than the reference set. For instance, the query set can be text, and the reference set can be videos (audio and images). Or, the query set can be text and the sketch of a time series, and the reference set can be the time series \cite{bamford2023multi}.

There are missing modalities of each instance due to sensor and storage failure or the infeasibility of modalities.
\textit{The missing modalities are not consistent across the same modality within the set.}
We define the observable reference set as $\obsdataset{\mathrm{r}}$ and the observable instances as $\obsdata{i}$. If an instance has missing modalities, we replace them with $\zero$, the zero vectors. Mathematically,
\begin{equation*}
    \obsdata{i} = (..., \data{i}{\modal{i_1}}, ..., \zero,.., \data{i}{\modal{i_2}}, \zero, ..., \data{i}{\modal{i_n}}, ...),
\end{equation*}
with $i_n$ as the number of observable modalities in instance $\obsdata{i}$.
The notation holds for the query set as well.

We want to design a similarity function $\newsim(\obsqdata{i}, \obsdata{j})$ that inputs the observable instances and replicates the original similarity of the complete modalities $\cossim(\qdata{i}{}, \data{j}{})$:
\begin{equation}
    \newsim = \argmin_{\forall \dot{\newsim}} \sum_{\forall i \in [1, \qnum] \\, j \in [1, \dpnum]} \| \dot{\newsim}(\obsqdata{i}, \obsdata{j}) - \cossim(\qdata{i}{}, \data{j}{})\|.
    \label{eq:sim2gt}
\end{equation}
We use $\dot{}$ to denote variables of optimization problems throughout the paper.
\Eqref{eq:sim2gt} is impossible to solve since we do not restrict the domain of the similarity function $S$ nor know the complete modalities. Hence, we can only solve \Eqref{eq:sim2gt} approximately.

Now, we describe the querying process in retrieval, which is our targeted task.
Given a query instance $\obsqdata{q}{}$, we want to retrieve the closest match in the reference set.
The definition of close depends on the task, and we can only use the similarity score as a proxy.
For text-to-video retrieval, we want to retrieve the video that corresponds to the text query. 
For place recognition, we want to retrieve the reference instance representing the closest location to the querying robot agent.
To do so, we calculate the query instance's similarity score to all instances in the reference set,  $\newsim(\obsqdata{q}, \obsdata{i}), ~ \forall i = 1,2,...,\dpnum$. 
We retrieve the most similar instance $\obsdata{\hat{q}}{}$ from the reference set, defined as:
\begin{equation}
    \hat{q} = \argmax_{\dot{q} \in [1, \dpnum]} \newsim(\obsqdata{q}, \obsdata{\dot{q}}).
    \label{eq:retrieval_results}
\end{equation}
The final metric of the retrieval performance is to see whether the closest match selected from $\hat{q}$ is correct or not, and we want to maximize this accuracy.

With the advances in cross-modal similarity metrics and encoders, the primary challenge in Any2Any retrieval is the incomplete modalities.
These missing modalities vary between the reference set and the query data, adding significant complexity to the process of calculating similarity scores since we cannot find a common modality among all instances and use it for retrieval.

%% file: section/preliminary.tex
\section{Conformal Prediction in Retrieval}
\label{sec:prelim}

We now extend conformal prediction to retrieval tasks.
Notably, the extended definition of conformal prediction presented here applies to all retrieval tasks, regardless of whether all modalities are available or not.

We first define a space of (similarity score, ground truth label) pairs as $\Theta \times \mathcal{Y}$.
The scores, \textit{e.g.}, CLIP cosine similarity scores, evaluate the similarity between a query instance and a candidate instance of retrieval. The ground truth labels are binary labels indicating whether the similarity scores lead to a correct retrieved instance or not.
Since the scores are bounded between $-1$ and $1$, we denote $\Theta=[-1, 1],~\mathcal{Y}=\{0, 1\}$.
We separately use a calibration set $\caliset = \{(\simscore_1, y_1),..., (\simscore_\calinum, y_\calinum)\}$ with size $\calinum$ from space $\Theta \times \mathcal{Y}$ to calibrate the similarity scores.
The calibration set consists of independent and identically distributed (i.i.d.) samples from a distribution over space $\Theta \times \mathcal{Y}$ and is disjoint from the train and test sets of any other models.

The goal is that given an error rate $\epsilon \in [0, 1]$, we want to find a set of labels such that this set covers the correct ground truth label with a probability greater than $1-\epsilon$.
We call the mapping of the similarity score to the subsets of labels \emph{prediction band} $\predband: \Theta \mapsto 2^{\mathcal{Y}}$ (power set of $\mathcal{Y}$).
We have the following assumption:

\textbf{Assumption 1:} During inference, a new pair $(\simscore_{\eta+1}, y_{\eta+1}) \in \Theta \times \mathcal{Y}$ is an i.i.d. sample from the distribution where we sample the calibration set $\Phi= \{(\simscore_1, y_1),..., (\simscore_\calinum, y_\calinum)\}$.

Given $\epsilon$, if assumption 1 holds, conformal prediction finds a prediction band $\predband$ that satisfies:
\begin{equation}
    1 - \epsilon + \frac{1}{\eta+1} \geq \mathbb{P}\left[y_{\eta+1} \in \predband(\simscore_{\eta+1};\epsilon) \right] \geq  1 - \epsilon.
    \label{eq:cp}
\end{equation}
We refer to $1 - \epsilon$ as the \textit{conformal probability}, representing the lower bound of correct retrieval. When $\eta$ is large enough, the upper bound is close to the lower bound, and we have more accurate estimations.
Now, we define the conformal retrieval problem with the conformal prediction:

\textbf{Problem 1 (Conformal Retrieval):}
Given a new similarity score $\simscore_{\eta+1}$, we want to find the maximum conformal probability $1 - \epsilon$ results in $\predband(\simscore_{\eta+1};\epsilon) = \{1\}$.
Namely, given a similarity score $\simscore_{\eta+1}$ between a query and reference instances, we are interested in guaranteeing with the probability $1 - \epsilon$ that it leads to a correct retrieval.

We show how to obtain the prediction band and omit the proof which is in \cite[Theorem D.1]{angelopoulos2021gentle}.
We first define the calibration scores as:
\begin{equation}
\begin{aligned}
    \caliscore_{i} = \caliscore & (\simscore_{i}, y_{i}) = \left \|y_{i} - \frac{\simscore_{i} - \min_j(\simscore_j)}{\max_j(\simscore_j) - \min_j(\simscore_j)} \right \|, \\
    & \forall i = 1,2,...,\eta. ~~~~ (\text{Calibration Score})
    \label{eq:cali_score}
\end{aligned}
\end{equation}
The calibration scores indicate the difference between the ground truth label and the predicted label from the similarity score, where higher similarity scores indicate $y_i=1$.
According to \cite[Section 1.1]{angelopoulos2021gentle}, the prediction band is:
\begin{equation}
\begin{aligned}
    \predband(\simscore;\epsilon) & = \{y ~|~ \caliscore(\simscore, y) \leq \alpha \}, \\
    & (\text{Inverse Solution to Conformal Retrieval})
    \label{eq:pred_band}
\end{aligned}
\end{equation}
where $\alpha$ is the $\frac{\ceil{(\eta + 1) (1-\epsilon)}}{\eta} \simeq (1-\epsilon)$ percentile of the calibration scores $\{ \caliscore_i ~ | ~ \forall i = 1,2,...,\eta \}$.
In \Eqref{eq:pred_band}, we actually solve the inverse of the conformal retrieval problem. We find the prediction band $ \predband$ given the conformal probability $1 - \epsilon$. To solve the conformal retrieval problem, we find: 
\begin{gather}
\begin{aligned}
    \epsilon = \inf_{0 \leq \dot{\epsilon} \leq 1} ~ \dot{\epsilon}, ~~~ & \text{s.t. } \predband(\simscore;\dot{\epsilon}) = \{y ~|~ \caliscore(\simscore, y) \leq \alpha \} = \{1\}. \\
    & (\text{Solution to Conformal Retrieval})
    \label{eq:cp_solution}
    \raisetag{20pt}
\end{aligned}
\end{gather}
Namely, we find the minimum acceptable error rate $\epsilon$ that results in correct retrieval, and the probability of correct retrieval is lower-bounded by $1 - \epsilon$.

\textbf{Remarks.}
One advantage of conformal prediction is that the methodology is independent of the machine learning model used to obtain the similarity score.
If the model is poor, the similarity score $\simscore$ does not correlate well with the retrieval result, leading to a looser estimated error rate $\epsilon$, and vice versa.
The choice of the calibration score function $\caliscore$ is essential for overall performance.

%% file: section/method.tex
\section{Any2Any Conformal Retrieval}
\label{sec:method}

\subsection{Cross-modal Similarity Scores}
We now explain the Any2Any framework in detail.
Per the related works section, one can use cross-modal encoders, such as CLIP, to calculate the similarity of instances.
We define the cross-modal similarity function of modalities $\qmodal{i}, \modal{j}$ as $\bisim^{\qmodal{i}, \modal{j}}$, which is a building block of our method.
We do not restrict $\qmodal{i} \neq \modal{j}$, so this definition includes unimodal similarity.
These functions do not input two instances directly, but two unimodal parts from the instances.

We now consider all combinations of cross-modal similarity scores of a query instance $\obsqdata{q}$ and a reference $\obsdata{i}$:
$$\bisim^{\qmodal{j}, \modal{k}}(\qdata{q}{\qmodal{j}}, \data{i}{\modal{k}}), ~~ \forall j,k ~~ \text{s.t. } \qdata{q}{\qmodal{j}}, \data{i}{\modal{k}}\neq \zero.$$
We denote the similarity matrix of instance pair $(\qdata{q}{\qmodal{j}}, \data{i}{\modal{k}})$ of which entries are the cross-modal similarity scores as:
\begin{equation}
\begin{aligned}
    \relax[\simmat{q}{i}]_{jk} = 
    \begin{cases}
    \bisim^{\qmodal{j}, \modal{k}}(\qdata{q}{\qmodal{j}}, \data{i}{\modal{k}}), & \text{if } \qdata{q}{\qmodal{j}}, \data{i}{\modal{k}}\neq \zero, \\
    -1, & \text{otherwise},
    \end{cases}
    \label{eq:sim_score}
\raisetag{20pt}
\end{aligned}
\end{equation}
where $[\cdot]_{jk}$ is the entry of a matrix at the $j$th column and $k$th row, and $\simmat{q}{i} \in \mathbb{R}^{\qmodnum \times \modnum}$.
Given the similarity matrices of two instance pairs $\simmat{q}{i}, \simmat{q}{j}$ with different modalities missing, how can we determine which pair is more similar in order to retrieve the more similar instance from $\obsdata{i}, \obsdata{j}$?
For example, we cannot directly compare:
\begin{equation}
\simmat{q}{i} = 
\begin{pmatrix}
    -1 & 0.7 \\
    -1 & -1
\end{pmatrix} \quad \gtreqless? \quad
\simmat{q}{j} = 
\begin{pmatrix}
    0.9 & -1 \\
    -1 & -1
\end{pmatrix}~.
\label{eq:example}
\end{equation}
The $0.7$ in $\simmat{q}{j}$ and $0.9$ from $\simmat{q}{i}$ are from two bimodal feature spaces, which are not comparable.

\subsection{Comparison of Similarity Matrices by Conformal Prediction.}
To address the incomparable issue, we use a two-stage conformal prediction to ground the similarity matrices.
The first stage grounds individual similarity scores in each cross-modal feature space to conformal probabilities.
The second stage fuses these conformal probabilities from a pair of query and retrieval instances to a scalar, enabling comparison across similarity matrices.

We convert all the entries in the matrices to their corresponding conformal probabilities, as described \Eqref{eq:cp}, where $[\simmat{q}{i}]_{jk}$ in \Eqref{eq:sim_score} is $\simscore$ in \Eqref{eq:cp}.
Note that similarity scores in each feature space are calibrated differently, thus we have $\eta\times\modnum$ prediction bands in total, denoted as $\predband^{\modal{j}, \modal{k}}$ for each cross-modal feature space of modalities $\modal{j}, \modal{k}$. 
They map the similarity scores in the space to the probabilities of correct retrieval.
We define the corresponding error rate as $\epsilon([\simmat{q}{i}]_{jk})$, which can be obtained by \Eqref{eq:cp_solution}.
Similarly to the similarity matrices, we can define conformal matrices $\conmat{}{}$ as:
\begin{equation}
    [\conmat{q}{i}]_{jk} = 1 - \epsilon([\simmat{q}{i}]_{jk}).
    \label{eq:conmat}
\end{equation}
The first stage of the conformal prediction is done. We ground the similarity scores from each feature space to conformal probabilities--the lower bounds of the probabilities of correct retrieval.
\input{figure_latex/modal_dist}
\Figref{fig:modal_dist} shows the incomparability of similarity scores from different modality encoders, Lip-loc \cite{shubodh2024liploc} and GTR \cite{ni2021gtr}, on the KITTI dataset \cite{KITTI}, detailed later in Sec. \ref{sec:exps}.
The two distributions of similarity scores lie in completely different ranges before calibration. For example, a score of $0.5$ represents the $55\%$-percentile in the LiDAR modality but the $0\%$-percentile in the text modality.
After the first stage calibration, the distributions both range between $[0, 1]$, thus in the same scale.

We still need a mechanism to enable direct comparison between the conformal matrices $\conmat{}{}$ in \Eqref{eq:conmat}.
We ground $\conmat{}{}$ to a statistically meaningful scalar by conformal prediction, which is the second stage.
We first use a mapping function $\mattonum: \mathbb{R}^{\qmodnum \times \modnum} \mapsto [0, 1]$ that maps the conformal matrices to a scalar.
For instance, it can be the mean or maximum of all non-$(-1)$ entries in the conformal probabilities matrices. 
Empirically, experiments show the mean function slightly better than the maximum function.
Again, we treat the output $\mattonum \left( \conmat{}{} \right)$ of the mapping function $\mattonum$ as $\simscore$ in \Eqref{eq:cp} to calibrate the conformal matrix to the conformal probability of correct retrieval with \Eqref{eq:cp_solution}:
\begin{gather}
\begin{aligned}
    \epsilon (\mattonum &(\conmat{q}{i}))  = \inf_{\dot{\epsilon}} ~ \dot{\epsilon}, \\
    \text{s.t. } & \predband(\mattonum \left( \conmat{q}{i} \right);\dot{\epsilon}) = \{y ~|~ \caliscore(\mattonum \left( \conmat{q}{i} \right), y) \leq \alpha \} = \{1\}.
\raisetag{30pt}
\end{aligned}
\end{gather}
Note that we use the same calibration set $\caliset$ for both stages of the conformal prediction.
Finally, we can compare the similarity scores from different cross-modal feature spaces, as exemplified in \Eqref{eq:example}.
To rewrite the retrieval results previously formulated in \Eqref{eq:retrieval_results} with our proposed method, we define the retrieved index given query instance $\obsdata{q}$ as:
\begin{equation}
    \hat{q} = \argmax_{\dot{q}\in [0, \dpnum]} ~ 1 - \epsilon(\mattonum(\conmat{q}{\dot{q}})).
    \label{eq:any_retrieval_results}
\end{equation}
We retrieve the instance from the reference set which has the maximum conformal probability with mapping function $\mattonum$ combining all multimodal similarity scores.

\input{figure_latex/calibration_plots}

The second stage of conformal retrieval effectively differentiates correct and incorrect retrieval distributions, as shown in \Figref{fig:calibration_dist}. The calibration score distributions for correct and incorrect retrievals reveal distinct patterns, emphasizing the method's ability to separate these distributions. \Figref{fig:calibration_curve} further shows that higher calibration scores correspond to an increased probability of correct retrieval.
This analysis validates the proposed method in a practical scenario, particularly since we retrieve only the top-$k$ similar instances, where $k$ is relatively small compared to the reference set.
Notably, unlike \Figref{fig:modal_dist} shows calibration scores on a specific modality, \Figref{fig:calibration} shows the overall scores on all modalities with some missing.

To sum up, the first stage grounds the similarity scores to probabilities in the same scale (\Figref{fig:modal_dist}), enabling the arithmetic operation of $\mattonum$ later used in the second stage (\Figref{fig:calibration}).
The second stage is necessary since we need a mechanism to fuse all entries in the conformal matrix $\conmat{}{}$ to a single probability to enable comparisons of scalars, not matrices.
The entire Any2Any retrieval framework is guaranteed by the error rate $\epsilon$ of the second stage with conformal prediction.

\subsection{Computation Overhead}
\label{sec:overhead}
For conventional retrieval tasks, the Faiss library \cite{douze2024faiss, johnson2019billion} enhances the efficiency of similarity searches and clustering of dense vectors.
For the Any2Any framework, one can also leverage Faiss to accelerate the retrieval process with a simple heuristic explained in the following.
To retrieve the top-$k$ similar instance, one first uses Faiss to retrieve the top-$\alpha k$ similar instances from all modalities individually, where $\alpha$ is an integer that balances the precision of the calculation and the retrieval.
Next, within the top-$\alpha k$ instances, we calculate the conformal probabilities using the prediction band (\Eqref{eq:cp_solution}). Solving \Eqref{eq:cp_solution} is equal to finding the percentile of the given similarity score within the set of calibration scores (\Eqref{eq:pred_band}).
If the calibration scores are sorted a priori, the time complexity to find the percentile is $O(\log |\Phi|)$ using binary search.
The overall computation overhead of Any2Any per query compared to conventional retrieval is: 
\begin{equation}
    O\left((\alpha - 1)k\qmodnum\modnum\log |\Phi| \right), ~~ \text{(Computation Overhead)}
    \label{eq:overhead}
\end{equation}
where the coefficient $\qmodnum\modnum$ is an upper bound of the number of observable entries in the similarity matrix $\simmat{q}{i}$.

As shown in \Eqref{eq:overhead}, the overhead scales linearly with $\alpha$. A larger $\alpha$ includes more similar candidates, increasing the likelihood that the top-$\alpha k$ instances contain the real instance with the maximum conformal probabilities across all modalities. However, it comes at the cost of increased computation.
In our experiments, we calculate all similar instances exhaustively to ensure the results are accurate.

%% file: figure_latex/modal_dist.tex
\begin{figure}[t!]
\centering
\begin{subfigure}[t]{0.35\textwidth}
    \centering
     \includegraphics[width=1\textwidth]{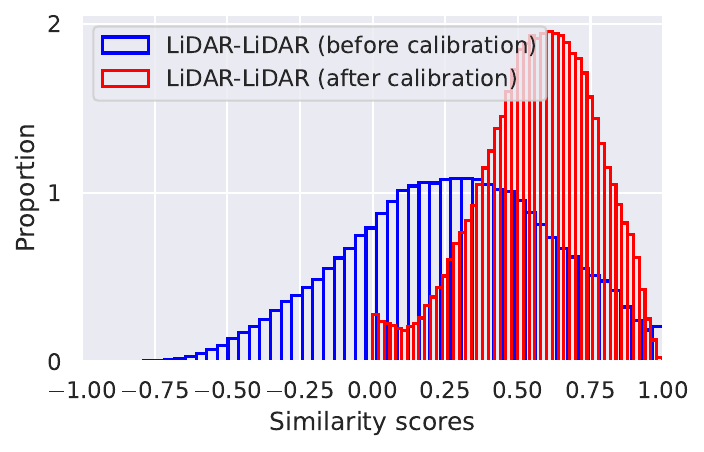}
    \caption{\small{LiDAR-to-LiDAR similarity scores.}}
    \label{fig:modal_dist_lidar}
\end{subfigure}
\begin{subfigure}[t]{0.35\textwidth}
    \centering
    \includegraphics[width=1\textwidth]{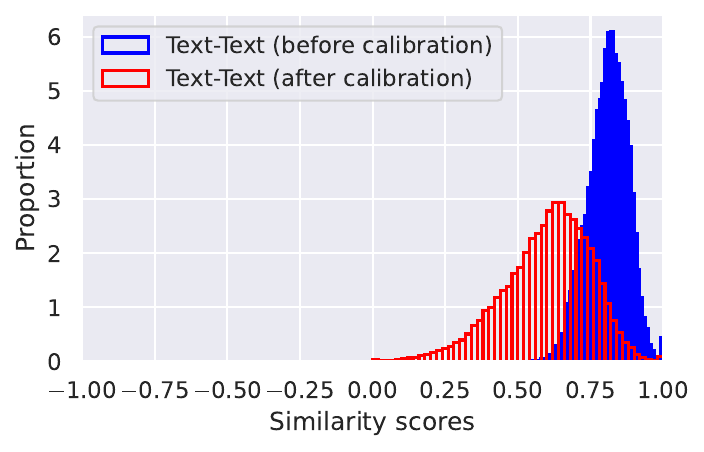}
    \caption{\small{Text-to-text similarity scores.}}
    \label{fig:modal_dist_text}
\end{subfigure}
  \caption{\small{\textbf{Normalizing similarity score distributions with first stage conformal prediction.}
  Comparison of similarity score distributions before (blue) and after (red) calibration with conformal prediction for (a) LiDAR-to-LiDAR and (b) text-to-text retrieval tasks in KITTI.
  Before calibration, the similarity scores of the two modalities fall in various ranges, and after calibration, they both range between $[0, 1]$, enabling direct comparison.
  }}
  \label{fig:modal_dist}
\vspace{-1em}
\end{figure}

%% file: figure_latex/calibration_plots.tex
\begin{figure}[t!]
\centering
\begin{subfigure}[t]{0.3\textwidth}
    \centering
     \includegraphics[width=1\textwidth]{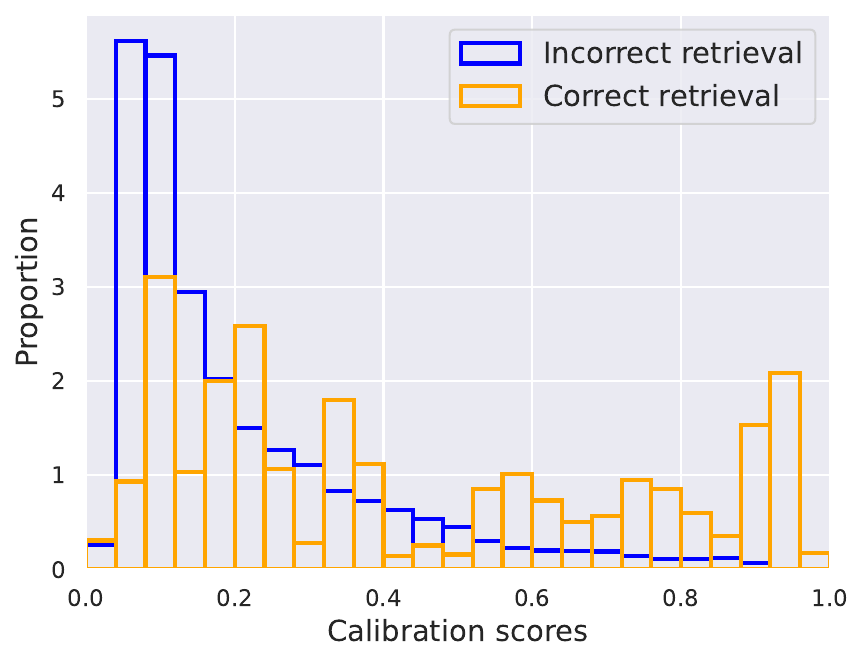}
    \caption{\small{Distribution of calibration scores.}}
    \label{fig:calibration_dist}
\end{subfigure}
\begin{subfigure}[t]{0.3\textwidth}
    \centering
    \includegraphics[width=1\textwidth]{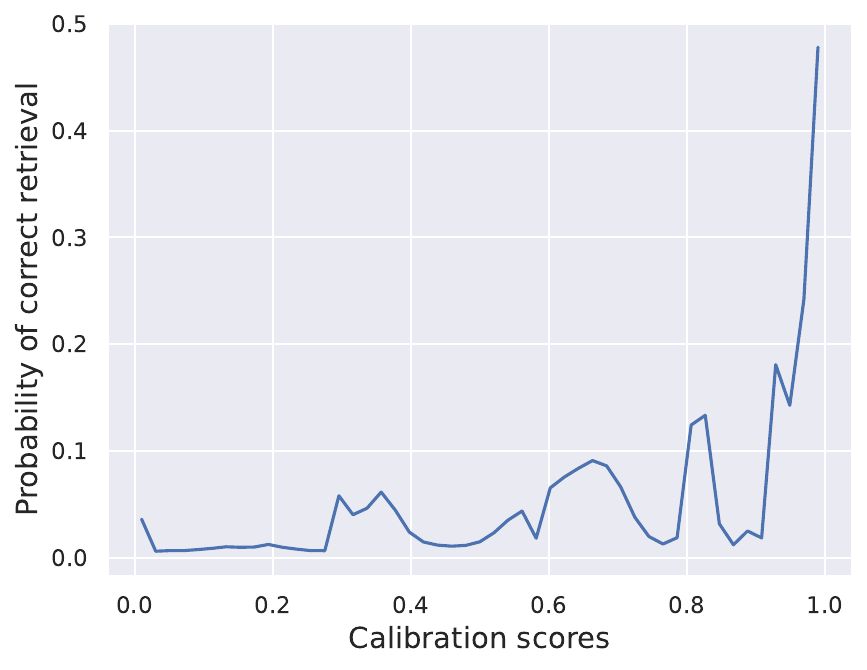}
    \caption{\small{Probability of correct retrieval.}}
    \label{fig:calibration_curve}
\end{subfigure}
  \caption{\small{\textbf{Conformal retrieval separates correct and incorrect distributions.}
  (a) shows the calibration score distributions for correct and incorrect retrievals, highlighting distinct patterns.
  (b) shows the probability of correct retrieval increases with calibration scores.
  We use the KITTI dataset's calibration set here.
}}
  \label{fig:calibration}
\vspace{-1.5em}
\end{figure}

%% file: section/experiments.tex
\section{Experiments}
\label{sec:exps}

\subsection{Datasets}
We tested the capability of the Any2Any framework on several multimodal retrieval datasets: (1) KITTI \cite{KITTI}, (2) MSR-VTT \cite{xu2016msr}, and (3) Monash Bitcoin \cite{bitcoin_dataset}.
Note that we trained the two-stage calibration process also on incomplete modalities here.
Refer to \Appref{app:exps_details} for more details on the experimental settings.

KITTI is a place recognition dataset for autonomous vehicles, containing RGB and LiDAR images from Karlsruhe, Germany. To incorporate the text modality, we use LLaVA \cite{liu2023improvedllava} to generate captions describing the images.
The task involves using the local RGB, LiDAR, and captions to help autonomous vehicles locate themselves within a city by identifying the closest landmark through retrieval. The landmarks are the reference set for retrieval, while local observations are the query data, making all data in this task trimodal ($\qmodnum=\modnum=3$).
We randomly remove entire modalities from either the query or reference instance with a probability of $50\%$.

MSR-VTT consists of YouTube video clips, and each clip corresponds to $20$ human-written annotations. We use the text to retrieve the video, which is naturally bimodal (image and audio), so $\qmodnum=1$ and $\modnum=2$.
Note that approximately $11\%$ of the videos are silent and lack audio, illustrating the setting with incomplete modalities.
We randomly remove the vision modality in the reference instances with a probability of $25\%$ and do not remove the audio modality since some are already missing, detailed in \Appref{app:exps_details}.
 
The Monash Bitcoin dataset consists of Bitcoin prices time series and daily summaries of news articles related to cryptocurrency.
Inspired by \cite{bamford2023multi}, We use the textual embeddings of the news summary to retrieve the actual time series.
The original dataset has two variants: missing data and non-missing data.
The missing data are due to technical issues or unavailability of information.
\textit{We are the first to perform paragraph-long textual retrieval on time series datasets, showing the Any2Any framework’s capability with previously unexplored modality pairs.}
We split the query instances into two parts: (1) historical news and (2) concurrent news. This division is made because the concatenated news embeddings have much larger dimensions than the time series embeddings ($1536$ vs. $120$), and splitting them allows for better similarity calculation. Additionally, since some news summaries are missing, having the concurrent news as a separate part ensures retrieval can still be performed using only the historical news.

The embeddings of time series are in two modalities: the original time domain and the extracted statistical features with tsfresh \cite{tsfresh}.
We incorporated statistical features since news articles commonly interpret time series through statistical features like maximum price and frequency of price change. 
Notably, while $\qmodnum=\modnum=2$, the query data modalities differ from those in the reference set, demonstrating the flexibility of Any2Any.
We randomly remove entire modalities from the query instances with a probability of $50\%$.

\input{figure_latex/single_modal}
\input{figure_latex/results}

\subsection{Baselines}
We show the retrieval performance of (1) Any2Any retrieval with incomplete modalities, (2) Any2Any retrieval with complete modalities (the Oracle baseline), and (3) each individual cross-modal retrieval.
We cannot compare with other baselines for retrieval with incomplete modalities \cite{shi2024incomplete, jing2020incomplete, li2022general}, as they require training new models, and no known architectures support all corss modalities, such as LiDAR and time series, in this context. Also, existing methods are limited to bimodal cases, whereas our approach accommodates any number of modalities.

\subsection{Results}
We show the performance of the Any2Any framework with and without complete modalities in \Tabref{table:results} and the performance of individual cross-modal retrieval in \Figref{fig:single_modal}. Results in \Figref{fig:single_modal} are not comparable baselines, but they give us more insight on the performance of Any2Any.

\textbf{KITTI.}
In \Figref{fig:single_modal}(a), all modality pairs achieve a similar performance with a recall rate of around $30\%$. It suggests that using LLaVA to generate captions for RGB images provides sufficient information about static objects in the image for place recognition, underscoring text as an effective modality for this task.
Any2Any matches the performance of the best individual modality pairs in \Figref{fig:single_modal}, as shown in \Tabref{table:results} for cases with and without incomplete modalities.

\textbf{MSR-VTT.}
Unlike \Figref{fig:single_modal}(b), the two modality pairs perform distinctly on MSR-VTT, as shown in \Figref{fig:single_modal}(b). 
Here, text-to-image retrieval achieves higher accuracy than text-to-audio retrieval, highlighting the lack of performant text-audio encoders. Any2Any's performance is slightly lower than using text-to-image retrieval alone because incorporating audio impacts overall results.
A simple but impossible heuristic can be: when the vision modality is available, retrieve with it; otherwise, use audio. Even with complete modalities, it is impossible due to the imcomparable issue between cross-modal similarities mentioned in \Secref{sec:method}.
Nonetheless, Any2Any outperforms this heuristic, whose Recall@$5$ is $37.6\%$ (three-forths of $49.3\%$ and one-forth of $2.6\%$), below Any2Any’s $42.5\%$ (\Tabref{table:results}).
It indicates that Any2Any is superior due to its two-stage calibration process, which not only selects the best modality but also integrates all available information in the case of incomplete modalities.
In summary, \textit{Any2Any's calibration process is more effective for scenarios with incomplete modalities than with complete ones.}

\textbf{Monash Bitcoin.}
\Figref{fig:single_modal}(c) shows that textual embeddings of news articles can be used to retrieve time series data.
We observe that text aligns more closely with statistical features, resulting in better performance. It supports the intuition to use them as a reference modality–news articles often interpret time series through statistical features.
The results of the Any2Any frameworks in \Tabref{table:results} are similar to the case in KITTI and MSR-VTT. Despite incomplete modalities, Any2Any can perform on par with the case of complete modalities in new and unexplored pairs of modalities.
The overall low performance is expected due to the novelty of these modality pairs, underscoring the need for better cross-modal similarity functions for text, trends, time series, and statistical features.
Notably, unlike other tasks, each query instance here has only one corresponding correct reference instance, so precision is not an ideal metric for evaluation.
Hence, we focus solely on recall.

\textbf{Summary of results.} The Any2Any framework with incomplete modalities matches the performance of the best individual modality pairs and the Oracle baseline with complete modalities in most cases, highlighting the effectiveness of the two-stage conformal prediction approach in compensating for incomplete modalities.
Note that \Figref{fig:single_modal} shows performance with complete modalities, making it incomparable to Any2Any. However, the performance of the complete Any2Any and best individual modality pairs establishes an upper bound of performance of the incomplete scenario. Also, the complete Any2Any sometimes performs worse than the incomplete case as the complete Any2Any is sensitive to low-performing modality pairs, which can negatively impact overall performance. In contrast, the incomplete case tends to be more robust, as the presence of missing data can mitigate the impact of weaker modalities.

\textbf{Can Any2Any generalize across encoders?}
In \Tabref{table:results}, we present the results for various encoders on MSR-VTT.
One experiment used ImageBind to encode both audio and images. The other used CLIP to encode images and CLAP \cite{laionclap2023} to encode audio. Both CLIP and CLAP encode text. 
\Tabref{table:results} shows that performance remains similar across different encoders on MSR-VTT, emphasizing that encoders serve as interchangeable components within the Any2Any framework. This compatibility allows the framework to function effectively with any encoder. One can always use state-of-the-art models, while they affect the upper bound of the overall performance.


\input{figure_latex/correlation}

\textbf{Do various modalities capture distinct information of the instances?}
A question that arises from the experiments is: are cross-modal similarity scores interchangeable? 
For instance, if we already have full image-LiDAR similarity scores, is there a need to evaluate text-LiDAR scores?
To address this question, we examined two correlations between the unimodal similarity scores: the canonical linear correlation and Spearman's rank correlation coefficient \cite{spearman1987proof}.
Spearman's rank correlation coefficient is a non-parametric measure that assesses the strength and direction of the monotonic relationship between two ranked variables, where values closer to $\pm1$ indicate stronger correlations.
As shown in \Figref{fig:corr}, there is no correlation between unimodal scores from the two encoders, suggesting that while all modalities yield similar performance in \Figref{fig:single_modal}, they capture distinct, uncorrelated information useful for retrieval.
Hence, cross-modal similarity scores are not interchangeable, as they capture distinct information from the instances. Any2Any offers a way to combine them harmoniously for retrieval.
We discuss how text and images capture different information in KITTI in \Appref{app:5m}.

\subsection{Limitations}
As discussed in \Secref{sec:overhead}, the Any2Any framework incurs additional computation overhead due to the prediction bands required by conformal prediction. We do not introduce any acceleration techniques specifically tailored for the Any2Any framework but only utilize Faiss.
Also, the calibration score function $\caliscore$ and the mapping function $\mattonum$ significantly impact the performance of Any2Any. Since the optimal choice remains unknown, we can only empirically test different functions.

%% file: figure_latex/single_modal.tex
\begin{figure*}[t!]
  \centering
\includegraphics[width=0.6\textwidth]{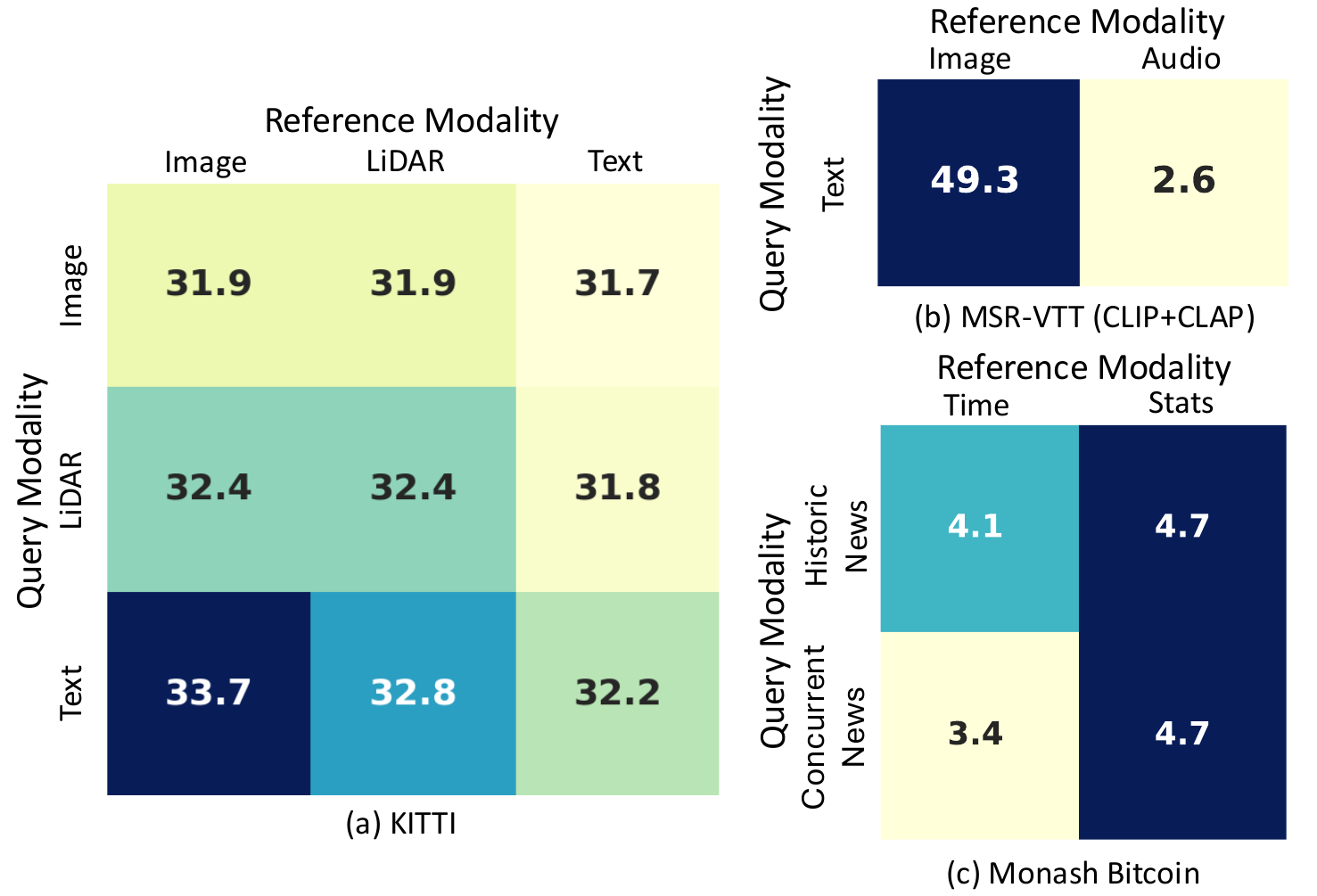}
\caption{\small{\textbf{Cross-modal retrieval performance with complete modalities.}
    The heatmaps show retrieval performance across various query and reference modality pairs on (a) KITTI, (b) MSR-VTT, and (c) Monash Bitcoin datasets. 
    Each cell represents the Recall@$5$ score, highlighting the variation in retrieval performance between modalities, with some modalities performing significantly better than others.
}}
  \label{fig:single_modal}
\end{figure*}

%% file: figure_latex/results.tex
\begin{table*}[t!]
    \centering
    \small
    \begin{tabular}{cccccccc}\hline
    Dataset & Any2Any Modality & mAP@$5$ & Precision@$1$ & Precision@$5$ & Precision@$20$ & Recall@$5$ & Recall@$20$ \\ \hline\hline
    \multirow{2}{*}{KITTI} & Incomplete & $4.5\%$ & $8.0\%$ & $9.0\%$ & $8.7\%$ & $\textbf{35.1}\%$ & $70.1\%$ \\
                            & Complete  & $4.0\%$ & $7.5\%$ & $7.9\%$ & $7.9\%$ & $\textbf{31.7}\%$ & $72.3\%$ \\ \hline
    MSR-VTT & Incomplete & $8.5\%$ & $26.8\%$ & $11.0\%$ & $3.7\%$ & $\textbf{42.5}\%$ & $55.0\%$ \\
    (CLIP+CLAP) & Complete  & $10.8\%$ & $30.5\%$ & $13.7\%$ & $4.8\%$ & $\textbf{48.3}\%$ & $64.0\%$ \\ \hline
    MSR-VTT & Incomplete & $14.8\%$ & $28.7\%$ & $17.3\%$ & $5.9\%$ & $\textbf{40.7}\%$ & $47.5\%$ \\
    (ImageBind) & Complete  & $18.9\%$ & $32.4\%$ & $22.6\%$ & $9.5\%$ & $\textbf{49.6}\%$ & $65.0\%$ \\ \hline
    \multirow{2}{*}{Monash Bitcoin} & Incomplete & $0.4\%$ & $0.6\%$ & $0.9\%$ & $0.7\%$ & $\textbf{4.7}\%$ & $15.5\%$ \\
                        & Complete  & $0.2\%$ & $0.0\%$ & $0.0\%$ & $0.7\%$ & $\textbf{3.3}\%$ & $15.5\%$ \\ \hline
    \end{tabular}
    \captionof{table}{\small{\textbf{Any2Any retrieval results.}
    The performance of Any2Any with incomplete modalities is comparable to that of Any2Any with complete modalities, demonstrating that the two-stage conformal prediction approach effectively compensates for incomplete modalities.
    We highlight Recall@$5$ for easier comparison with \Figref{fig:single_modal}.
}}
\label{table:results}
\vspace{-1em}
\end{table*}

%% file: figure_latex/correlation.tex
\begin{figure}[t!]
\centering
\begin{subfigure}[t]{0.23\textwidth}
    \centering
     \includegraphics[width=1\textwidth]{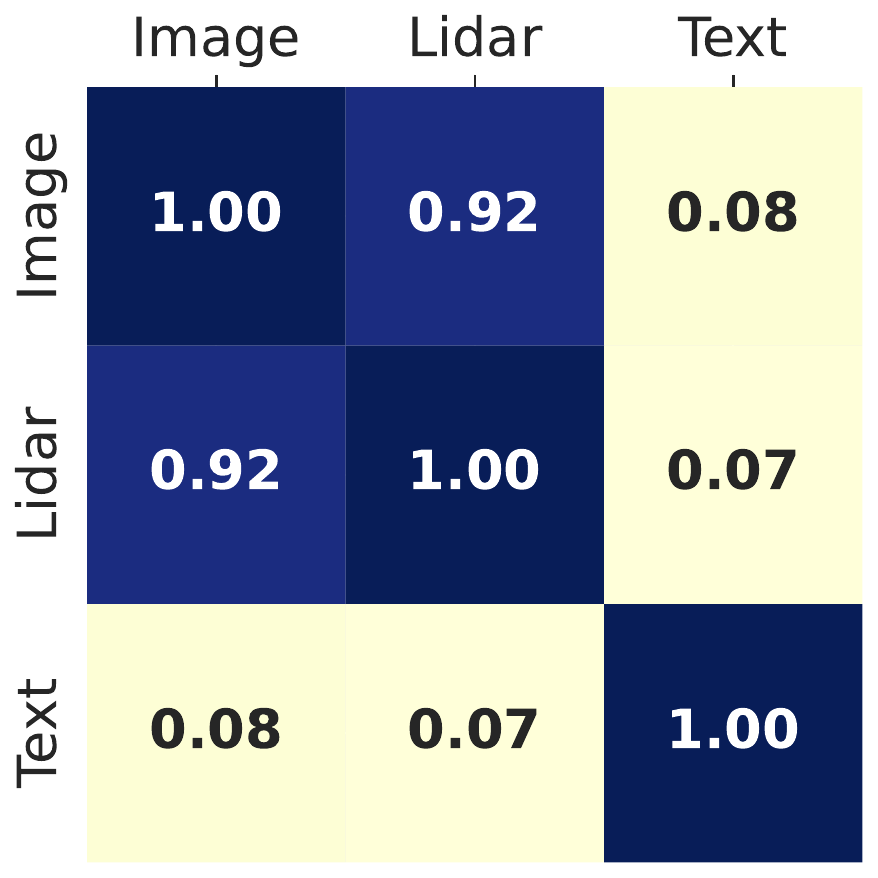}
    \caption{\small{Linear correlation.}}
    \label{fig:corr_}
\end{subfigure}
\begin{subfigure}[t]{0.23\textwidth}
    \centering
    \includegraphics[width=1\textwidth]{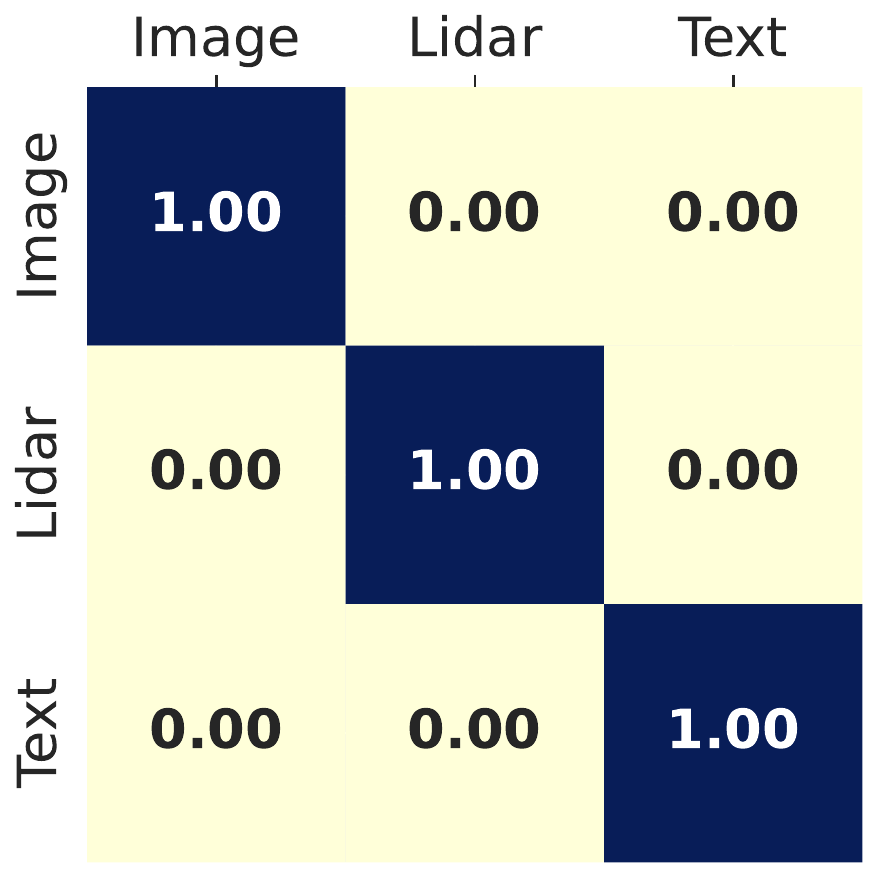}
    \caption{\small{Spearman’s rank correlation.}}
    \label{fig:corr_speaman}
\end{subfigure}
  \caption{\small{\textbf{Correlation coefficients between unimodal similarity scores.}
    There are no correlations between the unimodal scores from the two encoders, regardless of whether linear or Spearman correlations are considered. Lip-loc encodes LiDAR and image data, and GTR encoders text here.
}}
  \label{fig:corr}
\vspace{-1em}
\end{figure}

%% file: section/conclusion.tex
\section{Conclusion and Future Works}
In this work, we proposed a novel framework to retrieve multimodal data with varying incomplete modalities across instances.
We extended the definition of conformal prediction to retrieval and grounded the cross-modal similarities from various encoders to probabilities of correct retrieval. We thus enable direct comparison across retrieved instances.
We also showed that even if half of the instances have incomplete modalities, the Any2Any framework still achieves comparable performance to methods with complete information on multimodal retrieval tasks.

Previous works discussed compressing the vector embeddings to accelerate retrieval \cite{omama2024exploiting}, but compressing multimodal features \cite{li2023taskaware} for retrieval remains an open problem.
Lastly, an interesting direction for future work is to extend Any2Any by incorporating multiple cross-modal encoders, allowing the use of mixture of experts (MoE) approaches \cite{masoudnia2014mixture} or model selection algorithms \cite{li2024online, chen2023frugalgpt, timeseriesmodelselection} to combine and utilize the best-performing models.

%% file: section/appendix.tex
\section{Appendix}

\subsection{Additional Details on the Experiments}
\label{app:exps_details}
We run all inferences of encoder models and computation on an NVIDIA RTX A5000 GPU and a $64$ core Xeon Gold 6226R CPU machine. All experiments finished within $2$ hours from scratch since we do not involve any training of neural networks, only solving optimization problems.

\textbf{KITTI:}
We use Lip-loc \cite{shubodh2024liploc} to encode both image and LiDAR to the same embedding space and use cosine similarity to calculate the cross-modal similarity of these two modalities.
We use GTR \cite{ni2021gtr} to encode text and use cosine similarity to calculate the similarity of text embeddings.
We use CSA \cite{li2024csa} to map the two modalities to the same feature space and then use weighted cosine similarity to evaluate other cross-modal similarities, such as LiDAR-to-text.
The prompt for LLaVA to generate RGB image  captions is \textit{``Describe the static objects and the numbers of objects in the image within 20 words."}

The calibration set contains $1,097$ instances of multimodal data points, and the test set contains $6,000$ instances. We use the other $5,000$ instances to train CSA.
The retrieval task in the main paper is to locate the vehicle at any landmark within $20$ meters, per previous work \cite{shubodh2024liploc}. We decreased the distance to $5$ meters and increased the physical distance between the reference set.
We analyze the results in the next section.

\textbf{MSR-VTT:}
To obtain image and audio embeddings from raw videos, we divide each video into $2$-second clips and extract images and $2$-second audio segments from each clip. 
We use cosine similarity for all similarity evaluation here.
The calibration set contains $460$ human-labeled annotations, and the test set contains $5,000$ human-labeled annotations. We do not separate the reference set (the videos) into train and calibration sets. We use it as a whole and only separate the query set.

\textbf{Monash Bitcoin:}
Each time series contains hourly Bitcoin prices over $5$ days with a single channel, while the summary of news articles related to cryptocurrency is provided daily.
The historical news summaries represent the ones corresponding to the first $3$ days of the prices, and the concurrent ones corresponding to the last $2$ days.
We use OpenAI GPT-3.5 \cite{gpt3} to extract textual embeddings from the news articles and tsfresh \cite{tsfresh} to extract statistical features from the original time series. We remove the nan features due to the limited $120$ time steps horizon of the time series.
The train set to train CSA contains $524$ instances, the calibration set contains $71$ instances, and the test set contains $148$ instances.

\subsection{Coverage of Modalities in Place Recognition}
\label{app:5m}
\input{figure_latex/single_modal_5}
We showed the retrieval result on KITTI with a distance threshold of $5$ meters in \Figref{fig:single_modal_5}.
Compared to \Figref{fig:single_modal}(a), reducing the distance threshold for correct place recognition causes text to fail the task. It suggests that text captures global scene details but lacks accuracy in local details.
Interestingly, the cross-modal similarities exhibit asymmetric performance. For example, the text-to-LiDAR performance is significantly worse than LiDAR-to-text.
It is due to differences in modality specificity: text lacks the detailed query information needed to retrieve the correct instance but can effectively distinguish reference data, allowing it to filter out unrelated retrieval candidates.
Note that Lip-loc’s performance reaches nearly $100\%$ in this new setting, a significant improvement compared to \Figref{fig:single_modal}(a). It is due to the greater diversity in the reference set here when we sample from the videos in the original KITTI dataset, where the distances between retrieved candidates are larger, leading to more distinguishable similarity scores.

%% file: figure_latex/single_modal_5.tex
\begin{figure}[t!]
\centering
    \includegraphics[width=0.35\textwidth]{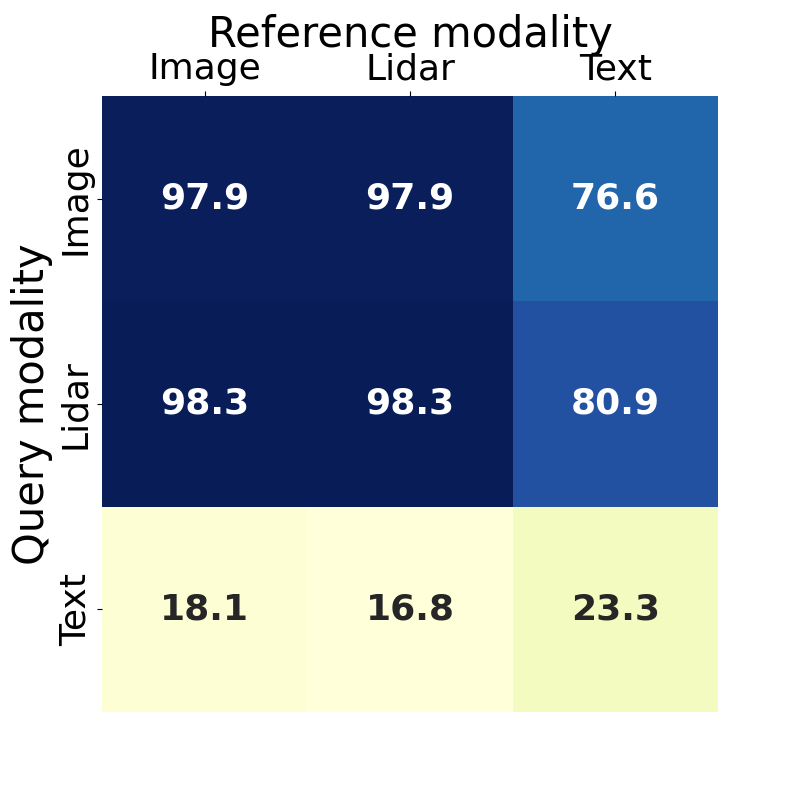}   
    \caption{\small{\textbf{Cross-modal retrieval performance on KITTI with a distance threshold of $5$ meters.} Compared to \Figref{fig:single_modal}(a), reducing the distance threshold for correct place recognition causes text to fail the task. It suggests that text captures global scene details but lacks accuracy in local details.
}} 
\label{fig:single_modal_5}
\end{figure}